\documentclass[letterpaper,10pt,conference]{ieeeconf}

\IEEEoverridecommandlockouts
\usepackage[utf8]{inputenc}
\usepackage[T1]{fontenc}
\usepackage{cite}
\usepackage{graphicx}
\usepackage{amsmath,amssymb}
\usepackage{booktabs}
\usepackage{array}
\usepackage{tabularx}
\usepackage{xcolor}
\usepackage{url}
\usepackage{tikz}
\usepackage{etoolbox}
\makeatletter
\apptocmd{\@xfloat}{\vspace*{5pt}}{}{\PackageWarning{root}{float pad patch failed}}
\pretocmd{\@makecaption}{\normalfont}{}{\PackageWarning{root}{caption font patch failed}}
\patchcmd{\@makecaption}{\scshape}{}{}{\PackageWarning{root}{caption small-caps patch failed}}
\makeatother
\usetikzlibrary{arrows.meta,positioning,fit,decorations.pathmorphing}
\DeclareRobustCommand{\halfcheck}{$\sqrt{}\mkern-9mu{\smallsetminus}$}
\DeclareRobustCommand{\fullcheck}{$\sqrt{}$}

\definecolor{draftred}{RGB}{170,35,35}
\definecolor{draftorange}{RGB}{190,100,0}
\definecolor{placeholdergray}{RGB}{245,245,245}

\newcommand{\method}{PreNav}
\newcolumntype{L}[1]{>{\raggedright\arraybackslash}p{#1}}
\newcolumntype{C}[1]{>{\centering\arraybackslash}p{#1}}
\newcolumntype{Y}{>{\centering\arraybackslash}X}
\newcommand{\tablenote}[1]{%
  \par\vspace{3pt}%
  \begin{minipage}{\columnwidth}
    \footnotesize\raggedright #1\par
  \end{minipage}}

\newcommand{\scene}[1]{\textsc{#1}}

\title{\LARGE \bf
Decide Before You Move: Diffusion-Generated Search Reference for \\
Zero-Shot Open-Vocabulary Object Navigation in \\ Semi-Static Environments
}

\author{
Yuxuan Feng$^{*}$, Fujing Xie$^{*}$, Sören Schwertfeger, and Hermann Blum
\thanks{$^{*}$Yuxuan Feng and Fujing Xie contributed equally to this work.}
\thanks{Yuxuan Feng, Fujing Xie, and Sören Schwertfeger are with the School of Information Science and Technology, ShanghaiTech University, Shanghai, China. Email: \{fengyx12025, xiefj, soerensch\}@shanghaitech.edu.cn.}
\thanks{Hermann Blum is with the University of Bonn and the Lamarr Institute, Bonn, Germany. Email: blumh@rpl.uni-bonn.de.}
\thanks{We provide our code and dataset at: \url{https://anonymous.4open.science/r/PreNav}}
}

\usepackage[hidelinks]{hyperref}

\begin{document}

% IEEEtran.bst control: "et al." beyond six authors, no dash for repeated authors (entry in root.bib)
\bstctlcite{IEEEexample:BSTcontrol}

\maketitle
\thispagestyle{empty}
\pagestyle{empty}

\begin{abstract}
Open-world navigation requires a robot to locate a user-specified target in an indoor scene under open-vocabulary queries.
Realistic indoor environments are \emph{semi-static}: walls and furniture largely persist while smaller objects move, relocate, or vanish between visits.
Prior-map systems answer queries from object poses recorded at reconstruction time, which expire as soon as objects move, whereas online methods re-derive where to look from current observations every few steps, often through expensive calls to a vision--language model.
Assuming that walls and large furniture do not move, we keep their map to plan paths, but use a target--furniture co-occurrence prior to decide where to search.
We present \method{}, an ObjectNav framework on semi-static environments.
Rather than invoking a vision-language model at each step, \method{} generates a single global search reference before the robot moves: an LLM converts target--furniture co-occurrence into a spatial prior, a conditional diffusion model turns that prior and the persistent structure into an ordered centerline with a band around it, and the robot approaches the target once it is detected online.
Experiments on a Gibson ObjectNav benchmark show that \method{} outperforms a strong online baseline under matched structural priors, using a generator trained only on room-level ScanNet++ scenes and applied to floor-level Gibson without fine-tuning. Real-world experiments under object relocation and open-vocabulary queries further validate the approach.
\end{abstract}

\section{Introduction}
Object navigation requires a robot to move through an indoor environment and find a target specified only by a user query~\cite{batra2020objectnav,anderson2018evaluation}. Existing methods keep either too much of the building or too little. Prior-map systems answer queries from dense object-level maps~\cite{werby2024hovsg,gu2023conceptgraphs}, whose object poses expire as soon as objects move. Map-free methods use no prior knowledge of the building and decide where to look online, re-deriving candidate regions from current observations every few steps, often through expensive calls to a vision--language model~\cite{yokoyama2024vlfm,padilla2026openfrontier}. Real buildings are \emph{semi-static}: walls and large furniture such as wardrobes and tables barely move, whereas objects such as keys and cups change position between visits. 

Between these extremes, two lines of work handle the semi-static building differently. One follows what changes: systems for changing scenes keep a semantic map of objects and update it online as they move, adding and removing voxels~\cite{liu2024dynamem} or tracking whether each object instance is still in place~\cite{bogenberger2026glasses}, yet their search still runs over the whole maintained map, movable objects included. The other builds on what persists: osmAG~\cite{feng2023osmag}, a lightweight, text-based hierarchical map of areas such as rooms and corridors, supports localization~\cite{xie2024agloc} and LLM-based map understanding and navigation~\cite{xie2024osmagcomp,xie2025copilot} even when incomplete, and osmAG-LLM~\cite{xie2026osmag} uses it to reliably identify the room that holds a moved or never-mapped target. Within that room, however, the robot can only check candidate locations recorded during mapping, namely viewpoints along the mapping trajectory and past object positions, which need not cover where the target now lies. With the search already narrowed to a room, and tracking every object as it moves requiring continual map maintenance, we ask whether the part of the room that barely changes, its walls and large furniture, suffices to plan where to look, and in what order, before the robot moves.

We argue that it does. Real buildings are semi-static: walls and large furniture persist across missions, while movable objects, including the target, may change location. Therefore, instead of continually updating object locations, or treating the environment as a completely unknown scene, we treat only the objects as unknown and rely on the layout of walls and furniture to infer where the target may be. Thus, even if the target has moved, we do not rely on previously recorded target locations. Instead, we use semantic scores between the target and furniture to guide the search: a cup may be near a dining table, whereas a television is unlikely to be near a toilet. 

We propose \method{}, which uses this prior knowledge of a scene to pre-plan an informed object search for a given query whenever a query arrives. An LLM scores target--furniture co-occurrence once, offline, and the scores are projected onto the mapped furniture as a spatial prior. A conditional diffusion model then turns this prior, the wall raster, and the start pose into what we call the \emph{search reference}: a globally planned reference path that optimizes the order of where to search in a floor or room. The reference is thus both an order to follow and a region to sweep. Trained on references from a solver for the traveling salesman problem (TSP), the generator produces one in about a second rather than tens of seconds. The robot then follows the reference, detects the target online, and approaches it, without repeated foundation-model calls or query-conditioned global replanning.

We evaluate \method{} on an expanded Gibson \cite{xia2018gibson} ObjectNav benchmark, asking three questions: whether it outperforms online baselines under matched structural priors; how generation efficiency trades against navigation quality when compared with its TSP teacher; and whether it transfers to real-world robots under open-vocabulary queries.

\begin{figure*}[!t]
  \centering
  \includegraphics[width=\textwidth,trim=13.8 187.6 10.0 181.6,clip]{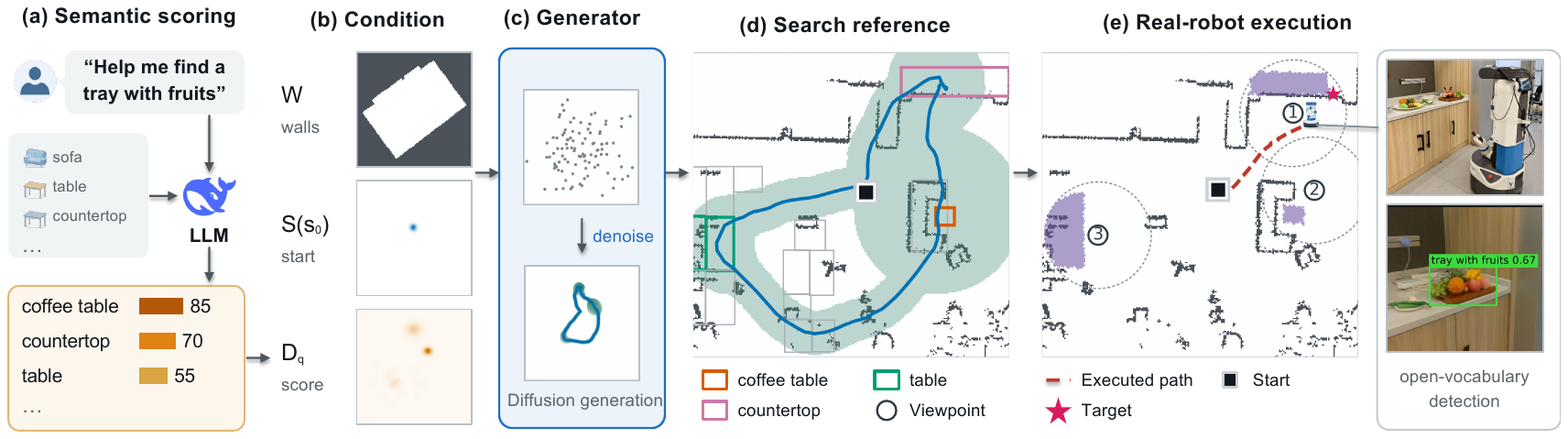}
  \caption{\textbf{Overview of \method{}.}
  (a)~For a query $q$, an LLM scores how useful each furniture category $c$ is for finding the target; the scores $r_c(q)$ are computed once and cached.
  (b)~The normalized weights $\alpha_c(q)$, projected onto the mapped furniture, form the spatial prior $D_q$, which is stacked with the wall raster $\mathcal{W}$ and the start map $S(s_0)$.
  (c)~Conditioned on these three channels, a diffusion model denoises random points into the search reference.
  (d)~The reference on the real lab map: a centerline that fixes the visiting order, and a band (teal) that widens over high-scoring furniture (outlined).
  (e)~Real-robot execution: viewpoints are sampled near the centerline to cover where the band overlaps furniture whose weight $\alpha_c(q)$ clears a threshold---the regions most worth searching, shown in violet. The robot visits them in order, detects the target with an open-vocabulary detector, and approaches it.
  Steps (a)--(d) run once, before the robot moves.}
  \label{fig:overview}
\end{figure*}

As a summary, our main contributions are:
\begin{itemize}
  \item We propose \method{}, an ObjectNav framework for semi-static environments: it guides online search through a search reference that is globally optimized with respect to the known room layout and semantic areas.
  \item We introduce an open-vocabulary interface that projects any user query onto the persistent furniture through LLM-generated co-occurrence scores between the query and each furniture category. As the generator's only query-dependent input, these scores let a conditional diffusion model plan a search reference for any query without retraining.
  \item We demonstrate zero-shot transfer from room-level ScanNet++ training to floor-level Gibson scenes, without fine-tuning the generator on Gibson. On the six-category benchmark, our method outperforms SGM under matched structural priors and achieves navigation performance comparable to its TSP teacher with an order of magnitude less generation time. We further demonstrate deployment on a real mobile robot with open-vocabulary queries.
\end{itemize}

\section{Related Work}

\subsection{Navigation with Prior Maps}

Prior-map systems organize a scanned building into persistent semantic representations and answer queries directly from this memory, without re-exploring. HOV-SG~\cite{werby2024hovsg} and ConceptGraphs~\cite{gu2023conceptgraphs} structure open-vocabulary 3D features into hierarchical scene graphs that support language queries over a pre-scanned building, and osmAG-LLM~\cite{xie2026osmag} tolerates such changes by searching online, but still draws its candidates from objects and views recorded during mapping. Their answers, however, depend on object poses recorded at reconstruction time---once objects move, the memory returns expired answers. \method{} likewise assumes a familiar building, but deliberately retains only walls and large-furniture regions; the current pose of the queried object is never provided.

\subsection{Navigation without Prior Maps}

Knowledge about the scene has to come from somewhere when no prior is given. Existing methods fall into three families. The first family decides on an \emph{online semantic map}: SemExp~\cite{chaplot2020semexp} predicts exploration goals on the map it builds, PONI~\cite{ramakrishnan2022poni} learns potential functions that prioritize promising regions, and SGM~\cite{zhang2024sgm} extends this map into unseen regions with a self-supervised generative model, selecting goals on the completed map. The second family explores through \emph{frontier scoring}: CoWs~\cite{gadre2023cows} scores frontiers with CLIP, VLFM~\cite{yokoyama2024vlfm} builds a vision--language value map, and OpenFrontier~\cite{padilla2026openfrontier} treats visually grounded frontiers as sparse semantic anchors. The third family follows \emph{route instructions} end-to-end: recent VLN methods such as NaVILA~\cite{cheng2025navila} fine-tune a vision--language model to map natural-language route descriptions and RGB observations to navigation actions.

The first two families require no persistent map, since the scene is entirely unknown when the task starts. They build a temporary map from observations and use the goal's semantics to decide the next action, and every few steps the robot re-decides through semantic reasoning, sometimes through expensive calls to a vision--language model. The third family builds no map; instead, each instruction already implies the scene it describes. The instruction itself is the prior---scene-specific, human-given. Of these, only the first family decides on a top-down semantic map into which known walls and furniture can be written before the episode starts; we therefore compare against its most recent member, SGM, under the same priors.

\subsection{Semi-Static Environments}

DynaMem~\cite{liu2024dynamem} adds and removes voxels as objects move, treating stable and movable parts of the scene alike. OpenIN~\cite{wang2025openin} and \emph{Where Did I Leave My Glasses?}~\cite{bogenberger2026glasses} separate the two, through static carriers or per-instance stationarity, and let an LLM rank where to search. All three, however, maintain a map that includes the movable objects and choose their next goal during navigation. \method{} keeps only walls and furniture, which need no maintenance, and plans the whole search before the robot moves.

\subsection{Diffusion Models for Planning}

Diffusion models learn to generate outputs from a training distribution by iterative denoising~\cite{ho2020ddpm}. In robot planning they serve at three levels: at the \emph{trajectory} level, Diffuser~\cite{janner2022diffuser} generates long-horizon state--action sequences; at the \emph{policy} level, NoMaD~\cite{sridhar2024nomad} navigates on topological memory with a single diffusion policy; and at the level of \emph{combinatorial solving}, TSPDiffuser~\cite{yonetani2025tspdiffuser} casts TSP as a denoising process over solution space, showing that diffusion can act as a learned sampler for combinatorial path planning. The last is closest to ours: the training labels likewise come from a combinatorial solver---TSP, the traveling salesman problem with neighborhoods, supplies the ground-truth path points---and we additionally learn a width at every path point, forming a band that marks where the target is likely to appear. This output is trained on room-level scenes, infers on floor-level scenes, and is validated on a real robot, with zero-shot and open-vocabulary queries supported by construction.

\section{Methodology}
\label{sec:method}

\method{} separates search-reference generation from online perception. When a query comes, the search reference is generated in two steps. First, the furniture-conditioned search prior (Sec.~\ref{sec:prior}) computes a spatial prior that encodes where the target is likely to be found relative to known furniture. Second, the diffusion-based global search generator (Sec.~\ref{sec:diffusion}) takes this prior, the persistent wall raster, and the start pose as inputs and generates a single search reference. The robot follows this reference, detects the target online, and then approaches it; subsequent execution requires neither repeated high-level foundation-model calls nor query-conditioned global replanning. Figure~\ref{fig:overview} summarizes the complete data flow.

\subsection{Furniture-Conditioned Search Prior}
\label{sec:prior}

In a semi-static environment, the target's current location is unknown, but its co-occurrence with persistent furniture provides a stable search prior. For example, a book is likely near a bookshelf, whereas an earphone is unlikely to appear near a refrigerator. We capture this through a two-step process: an LLM scores target--furniture co-occurrence, and the resulting weights are projected onto the spatial layout of the mapped furniture.

\textbf{LLM co-occurrence scoring.}
Let $\mathcal{C}$ denote the vocabulary of the supplied furniture prior. A query $q$ is free-form text that describes the target: an object category such as \emph{vase}, a description such as \emph{tray with fruits}, or a broader concept such as \emph{computer parts}. For every persistent furniture category $c\in\mathcal{C}$, an LLM is asked how useful $c$ is for finding the target under each of three spatial relations---the target resting \emph{on} $c$, contained \emph{in} $c$, or merely \emph{near} $c$---and returns the three relation scores together with a single search-usefulness score $r_c(q)\in[0,100]$ that summarizes them. Only $r_c(q)$ enters the prior. We normalize it per query to obtain weights,
\begin{equation}
  \alpha_c(q) =
  \frac{\exp\!\big(r_c(q)/\tau_q\big)}
       {\sum_{c'\in\mathcal{C}}\exp\!\big(r_{c'}(q)/\tau_q\big)},
  \qquad \tau_q = 0.65\,\sigma_q,
  \label{eq:softmax}
\end{equation}
where $\sigma_q$ is the standard deviation of $\{r_c(q)\}_{c\in\mathcal{C}}$. The temperature is set per query rather than globally because the LLM does not use a consistent range across queries: on the Gibson queries the top score is 95 for \emph{chair} but 50 for \emph{oven}, and a single temperature would leave the conservatively scored queries with a near-uniform prior. Scaling by the spread makes $\alpha_c(q)$ depend on how sharply the LLM separated the supports, not on the absolute numbers it happened to emit. We omit furniture categories with $\alpha_c(q)<0.01$ when constructing the condition map. These scores are computed once per query and cached; a new query requires one run of the same scoring utility before the robot moves, rather than LLM calls during navigation.

\textbf{Spatial prior construction.}
Each furniture instance $f_i$ with category $c(i)$ is reduced to its extent on a 2D $64\times64$ raster plane. It induces an axis-aligned anisotropic Gaussian $K_i(y,x)$ centered at the box centroid, whose per-axis standard deviations are half the corresponding box extents, each lower-bounded at $1.5$ cells so that small supports remain visible to the planning grid. The query-conditioned spatial prior is
\begin{equation}
  D_q(y,x) =
  \frac{\sum_i \alpha_{c(i)}(q)K_i(y,x)}
       {\sum_{y',x'}\sum_i \alpha_{c(i)}(q)K_i(y',x')} .
  \label{eq:distribution}
\end{equation}
$D_q$ is a spatial weighting over furniture supports rather than a target-occupancy prediction: the label generator seeks to cover the retained supports, while $D_q$ conditions the learned route and sets its supervised search-band width through Eq.~\eqref{eq:width}---for instance, a query for \emph{vase} gives wider width targets near dining tables. Because the weights are assigned to furniture categories rather than instances, they remain applicable when an object relocates between compatible supports.

\subsection{Diffusion-Based Global Search Generation}
\label{sec:diffusion}

The search reference must capture long-range dependencies, such as which supports to visit and in what order, while remaining flexible to local geometric constraints. Diffusion models are well-suited for this task because they learn multimodal distributions over structured outputs through iterative denoising, avoiding the combinatorial complexity of explicit tour optimization at inference.

\textbf{Problem formulation.}
The model receives a three-channel condition
\begin{equation}
  \mathbf{c}=[\,\mathcal{W},\; S(s_0),\; D_q\,],
  \label{eq:condition}
\end{equation}
where $\mathcal{W}$ is the wall/free-space raster, $S(s_0)$ is a Gaussian centered at the start pose with a standard deviation of 1.25 cells, and $D_q$ is the spatial prior from Eq.~\eqref{eq:distribution}. It denoises a sequence $\mathbf{G}\in\mathbb{R}^{L\times3}$ whose rows are $(y_j,x_j,w_j)$ targets: $(y_j,x_j)$ is a centerline point on the planning grid and $w_j$ specifies the desired search half-width around that point. We refer to $\mathbf{G}$ as the search reference: its centerline gives the visiting order, and its half-widths define the band swept around that centerline.

\textbf{Training data and reference supervision.}
\label{sec:supervision}
We generate structured reference labels offline from ScanNet++~\cite{yeshwanth2023scannetpp} rather than training from embodied rollouts. Each training example contains a wall raster, persistent furniture instances, a start location, a target category, and an ordered search reference. The dataset contains 83 eligible scenes, split 75/8 into training and validation; Gibson \cite{xia2018gibson} is reserved for evaluation and never used during training. For each empirical support-frequency vector, the data generator adds up to five randomly sampled support-weight variants, exposing the denoiser to diverse spatial priors rather than only the fixed co-occurrence table used at evaluation. During training, horizontal and vertical flips and rotations by multiples of $90^\circ$ are applied consistently to all condition channels and reference coordinates.

To produce the labels, we follow TSPDiffuser~\cite{yonetani2025tspdiffuser} and solve a TSP with neighborhoods~\cite{arkin1994tspn} offline: around the start and each furniture center we take a free-space neighborhood, and CP-SAT~\cite{perron2023cpsat} selects one point per neighborhood and a closed visit order under A* path costs~\cite{hart1968astar}. Joining the selected points by their A* paths gives a loop from the start, which we cubically interpolate to $L=100$ samples clamped to the planning grid. At each sample, the width target is derived from the same spatial prior:
\begin{equation}
  \begin{aligned}
    w_j &= w_{\min}+(w_{\max}-w_{\min})D_q(y_j,x_j),\\
    \mathbf{g}_j&=(y_j,x_j,w_j).
  \end{aligned}
  \label{eq:width}
\end{equation}
Here $w_{\min}=1$ and $w_{\max}=12$ grid cells (half-width). All three output channels and all condition channels are normalized to $[-1,1]$ during training. TSP is used only to create these offline labels and is absent at inference.

\textbf{Training objective.}
Let $\mathbf{G}_0$ be such a TSP label, and $\mathbf{G}_k$ its noised version at diffusion timestep $k$, drawn uniformly from $0\le k<1000$ for each sample. The standard noise-prediction loss is
\begin{equation}
  \mathcal{L}_{\mathrm{diff}} =
  \mathbb{E}_{\mathbf{G}_0,k,\epsilon}
  \left[\left\|\epsilon-
  \epsilon_\theta(\mathbf{G}_k,k,\mathbf{c})\right\|_2^2\right].
  \label{eq:diffusion}
\end{equation}
We additionally encourage the predicted centerline to avoid walls. For low-noise steps ($k<200$), we reconstruct the denoised estimate $\widehat{\mathbf G}_0=(\mathbf G_k-\sqrt{1-\bar\alpha_k}\,\epsilon_\theta)/\sqrt{\bar\alpha_k}$ and bilinearly sample the wall channel at the predicted centerline points. For a minibatch of $N$ samples, the implementation averages this penalty only over samples whose timestep satisfies $k_i<200$:
\begin{equation}
  \begin{aligned}
  \mathcal L_{\mathrm{wall}}
  &=\frac{\displaystyle\sum_{i=1}^{N}\mathbf{1}[k_i<200]
  \left(\frac{1}{L}\sum_{j=1}^{L}
  B_{\mathcal W}^{(i)}(\widehat y_{ij},\widehat x_{ij})\right)}
  {\displaystyle\max\!\left(1,\sum_{i=1}^{N}\mathbf{1}[k_i<200]\right)},\\
  \mathcal L&=\mathcal L_{\mathrm{diff}}+0.2\,\mathcal L_{\mathrm{wall}},
  \end{aligned}
  \label{eq:training_loss}
\end{equation}
where $B_{\mathcal W}^{(i)}(y,x)\in[0,1]$ is the sampled wall value for sample $i$. Thus, the auxiliary term is applied only to low-noise training timesteps (the late portion of the reverse process), where reconstruction of $\mathbf G_0$ from the predicted noise is numerically better conditioned.

\textbf{Architecture.}
A convolutional encoder maps the $64\times64$ condition to a 768-channel $16\times16$ feature grid; fixed 2D sinusoidal positions are added before the 256 map tokens are projected to $d=256$. Each three-dimensional reference token is projected to the same width and receives a fixed 1D sinusoidal position and a learned timestep embedding. A four-layer pre-norm Transformer decoder (eight heads, 1,024-dimensional feed-forward layers) applies self-attention over the reference tokens and cross-attention to the map tokens, then predicts one three-dimensional noise vector per reference token. The network has 5.24M trainable parameters and uses 1,000 squared-cosine diffusion steps.

\textbf{Inference.}
At deployment, the diffusion model uses the same channel order and normalization as during training. Starting from Gaussian noise, it produces one 100-point search reference in 50-step DDPM sampling. This is the only query-conditioned global decision made for the entire deployment.

\section{Experiments}
\label{sec:experiments}

We evaluate whether persistent structural priors improve ObjectNav, how diffusion trades navigation performance against the generation cost of its TSP teacher, and whether the framework transfers to real-world object queries. The main comparison and subsequent analyses use a six-category subset; we also report the complete ten-category evaluation.

% Queue the paired comparison tables here for the next page, alongside Baseline Comparison.
\begin{table*}[!t]
\centering
\begin{minipage}[t]{\columnwidth}
\centering
\small
\caption{ObjectNav on the six-category Gibson subset (600 episodes)}
\label{tab:main}
\begin{tabular}{@{}L{1.9cm} C{0.75cm} C{1.20cm} C{0.95cm} C{1.0cm}@{}}
\toprule
Method & Wall & Furniture & SR \(\uparrow\) & SPL \(\uparrow\) \\
\midrule
SGM & $\times$ & $\times$ & 55.0 & 29.6 \\
 & $\times$ & \halfcheck & 60.5 & 33.1 \\
 & \fullcheck & $\times$ & 61.3 & 36.2 \\
 & \fullcheck & \halfcheck & 62.0 & 36.2 \\
\midrule
Ours & \fullcheck & \halfcheck & 64.3 & 39.2 \\
 & \fullcheck & \fullcheck & \textbf{70.3} & \textbf{39.8} \\
\bottomrule
\end{tabular}
\tablenote{$\times$: no prior; \fullcheck: wall or full-scene furniture prior; \halfcheck: furniture categories supported by SGM.}
\end{minipage}\hfill
\begin{minipage}[t]{\columnwidth}
\centering
\small
\caption{ObjectNav on the full ten-category Gibson set (1,000 episodes)}
\label{tab:main_tencat}
\begin{tabular}{@{}L{1.9cm} C{0.75cm} C{1.20cm} C{0.95cm} C{1.0cm}@{}}
\toprule
Method & Wall & Furniture & SR \(\uparrow\) & SPL \(\uparrow\) \\
\midrule
SGM & $\times$ & $\times$ & 36.4 & 22.0 \\
 & $\times$ & \halfcheck & 40.6 & 25.5 \\
 & \fullcheck & $\times$ & 39.9 & 24.1 \\
 & \fullcheck & \halfcheck & 42.8 & \textbf{26.9} \\
\midrule
Ours & \fullcheck & \halfcheck & 44.2 & \textbf{26.9} \\
 & \fullcheck & \fullcheck & \textbf{44.5} & 26.1 \\
\bottomrule
\end{tabular}
\tablenote{$\times$: no prior; \fullcheck: wall or full-scene furniture prior; \halfcheck: furniture categories supported by SGM.}
\end{minipage}
\end{table*}

\subsection{Experimental Setup}

\textbf{Datasets and benchmark.}
We evaluate on the five validation scenes of Gibson~\cite{xia2018gibson} v1.1: \scene{Collierville}, \scene{Corozal}, \scene{Darden}, \scene{Markleeville} and \scene{Wiconisco}. Of the 15 annotated semantic categories, five serve as persistent furniture and the other ten as query targets, with 100 episodes each, 1,000 in total. Four of these targets, \emph{book}, \emph{bottle}, \emph{clock} and \emph{cup}, are limited by detection rather than by search; the main comparison and subsequent analyses therefore use the remaining six---\emph{chair}, \emph{oven}, \emph{potted plant}, \emph{refrigerator}, \emph{tv} and \emph{vase}---comprising 600 episodes. Sec.~\ref{sec:tencat} details the detector analysis behind this choice and reports the full ten-category results.

\textbf{Comparison protocol.}
Within each evaluation set, all methods use the same episodes, a 500-action budget and the ObjectNav success criterion used by SGM~\cite{zhang2024sgm} and PONI~\cite{ramakrishnan2022poni}. We report success rate (SR) and success weighted by path length (SPL)~\cite{chaplot2020semexp,ramakrishnan2022poni}, both as percentages. All ablation experiments use Mask R-CNN as the object detector, the same detector used by the baselines. To separate search from recognition, we further evaluate \method{} with stronger detectors and with a \emph{3DSG oracle}. Gibson's object labels in Habitat come from the 3D Scene Graph (3DSG) annotation~\cite{armeni2019_3dsg}, which Habitat converts into a semantic mesh and renders through its semantic sensor; the 3DSG oracle replaces the detector with these rendered labels, which cover every object pixel in the current frame. SGM's released code provides the same oracle. Success, in contrast, is judged on the top-down GT semantic map of the ObjectNav dataset~\cite{chaplot2020semexp}, a separate annotation that does not fully coincide with 3DSG (Sec.~\ref{sec:tencat}).

SGM's pretrained semantic model supports the dataset's original 15 semantic categories, including five that we treat as furniture: \emph{dining table}, \emph{couch}, \emph{bed}, \emph{toilet} and \emph{sink}. For a fair comparison, we supply both SGM and \method{} with the same wall prior and restrict their furniture priors to these five categories (\halfcheck in Tables~\ref{tab:main} and~\ref{tab:main_tencat}). \method{} also accepts open-vocabulary furniture categories through its spatial-prior interface, without retraining the generator. We therefore additionally evaluate it with the full-scene furniture prior (\fullcheck), which adds the furniture that the scenes leave unlabeled, annotated by us as 2D boxes; this is a comparison within \method{} to measure the benefit of additional furniture information.

\textbf{Implementation details.}
The denoiser is trained for 800 epochs using AdamW at $10^{-4}$, a cosine learning-rate schedule with 100 warm-up steps, gradient clipping at 1.0, and train/validation batch sizes of 32/8. We evaluate in Habitat~\cite{savva2019habitat} under the common action-budget protocol. The agent carries a $640\times480$ RGB-D sensor ($79^\circ$ HFOV, 0.88\,m height) with discrete actions (0.25\,m forward and $30^\circ$ turns) and a 5\,cm occupancy grid. Category-specific perception uses Mask R-CNN R-50-FPN-3x~\cite{he2017maskrcnn} trained on COCO~\cite{lin2014coco}; score thresholds are 0.3 (ten-category) and 0.7 (six-category), applied identically to all methods. The co-occurrence scores are produced once by DeepSeek V4 Pro and normalized with the per-query softmax of Eq.~\eqref{eq:softmax}. The diffusion model samples one 100-point search reference in 50 denoising steps, with the half-width channel clipped to $[1,12]$ grid cells. Diffusion and TSP receive identical semantic inputs.  All experiments run on an NVIDIA RTX 4090 24GB; generator timing is measured on an AMD Ryzen 9 9950X with 16 single-thread workers (Table~\ref{tab:tsp_diffusion}).

\subsection{Baseline Comparison}

\textbf{SGM benefits from structural priors.}
The walls and furniture we supply to SGM are taken from the top-down GT semantic map. SGM is itself trained to complete such maps from partially masked ones, so giving it part of the GT map is exactly the input its training prepares it for. Furniture alone raises its SR by 5.5\% and SPL by 3.5\%, and walls alone by 6.3\% and 6.6\% (Table~\ref{tab:main}); with both, its SR and SPL rises by 7.0\% and 6.6\% over the original SGM. These priors are therefore informative, and SGM exploits them as its training intends.

\textbf{\method{} exploits the same priors better, and gains from more furniture.}
SGM receives furniture as a pixel-level semantic map, whereas \method{} needs only a 2D bounding box per piece (Sec.~\ref{sec:prior}), a coarser description. Even so, with the same walls and five furniture categories, \method{} outperforms SGM by 2.3\% in SR and 3.0\% in SPL. To show that furniture drives this search, we additionally annotate, as 2D boxes, the furniture that the scenes leave unlabeled; it lies outside the 15 annotated categories, and hence outside what SGM's semantic model supports. This full-scene prior raises the SR of \method{} by another 6.0\% and its SPL by 0.6\%. \method{} thus exploits both the layout and the semantics of walls and furniture, and since each additional piece costs only a 2D box, supplying more furniture is a cheap way to make the target more likely to be found.

\subsection{The Full Ten-Category Benchmark}
\label{sec:tencat}

\begin{figure}[!t]
  \centering
  \includegraphics[width=\columnwidth]{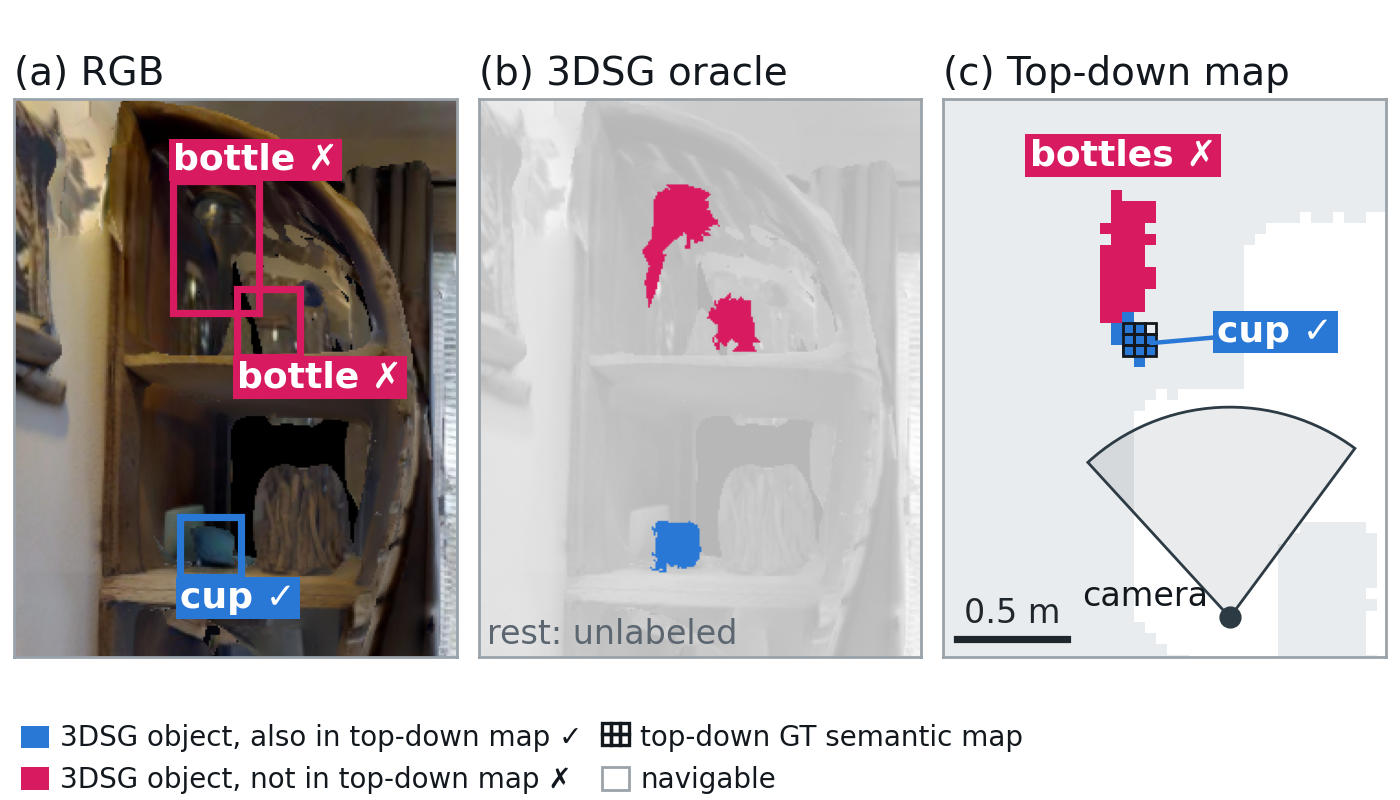}
  \caption{The 3DSG oracle and the success criterion read different annotations (\scene{Collierville}). (a)~RGB view of a shelf, with each 3DSG object boxed. (b)~The 3DSG oracle labels two bottles and a cup. (c)~The top-down GT semantic map used to judge success marks the cup (grid) but has no bottle anywhere in the scene; 3DSG objects are projected onto it for comparison, and the wedge marks the camera's field of view.}
  \label{fig:annot_mismatch}
\end{figure}

\textbf{Why the main analysis uses six categories.}
The Gibson ObjectNav data annotate 15 semantic categories, but the validation set queries only six of them, and three of these---\emph{couch}, \emph{bed} and \emph{toilet}---are furniture in our setting, leaving only three targets. We therefore extend the query set: of the 15 categories, five serve as furniture and the other ten as targets, with 100 episodes each, 1,000 in total. Detector ablation (Table~\ref{tab:tencat}) shows that four of the added targets are limited by perception rather than by search. On \emph{book}, \emph{bottle}, \emph{clock} and \emph{cup}, \method{} succeeds in only 21.2\% of episodes with Mask R-CNN; the stronger SAM~3, which improves the other six categories by 3.8\%, lowers them by 9.4\%, and even the 3DSG oracle raises them only to 47.0\%. Part of this gap comes from the annotations themselves: the oracle and the success criterion read different annotations, which disagree even at the category level. In the validation scene \scene{Collierville}, the 3DSG annotation contains two bottles that the top-down GT semantic map lacks, although both annotations mark the cup on the shelf below (Fig.~\ref{fig:annot_mismatch}). Conversely, in \scene{Wiconisco} the map contains a clock while the scene's 3DSG annotation has no \emph{clock} class, so the oracle can never report the target in the 57 \emph{clock} episodes set there. This is why Mask R-CNN exceeds the 3DSG oracle on \emph{clock} in Table~\ref{tab:tencat} (34\% vs.\ 28\%): in \scene{Wiconisco} it succeeds in 32 of these 57 episodes, the oracle in none. To measure navigation rather than recognition while maintaining comparability with the Object-Nav benchmark, we report the remaining six categories as the main benchmark and the full ten categories here.

\textbf{Results on all ten categories.}
The same priors again help SGM (Table~\ref{tab:main_tencat}): furniture alone raises its SR by 4.2\% and SPL by 3.5\%, walls alone by 3.5\% and 2.1\%, and both together by 6.4\% and 4.9\%. Under these priors, \method{} outperforms SGM by 1.4\% in SR with equal SPL, and the full-scene furniture prior adds a further 0.3\% in SR, at 0.8\% lower SPL. This slightly lower SPL indicates that the additional prior information leads to a few more found objects, but at the expense of generally slightly longer routes.
\begin{table}[!t]
\centering
\small
\caption{Comparison of per-category success rates (\%) across different detectors}
\label{tab:tencat}
\setlength{\tabcolsep}{4pt}
\begin{tabular}{@{}lccc@{}}
\toprule
Category & Mask R-CNN & SAM~3 & 3DSG oracle \\
\midrule
chair & 70 & \textbf{73} & 98 \\
oven & 79 & \textbf{91} & 96 \\
potted plant & \textbf{80} & 54 & 81 \\
refrigerator & 41 & \textbf{71} & 75 \\
tv & 54 & \textbf{57} & 74 \\
vase & 33 & \textbf{34} & 66 \\
\midrule
\textit{book} & \textbf{29} & 18 & 53 \\
\textit{bottle} & \textbf{20} & 16 & 67 \\
\textit{clock} & \textbf{34} & 12 & 28 \\
\textit{cup} & \textbf{2} & 1 & 40 \\
\midrule
Six-category subset & 59.5 & \textbf{63.3} & 81.7 \\
\textit{Other four} & \textbf{21.2} & 11.8 & 47.0 \\
All 1,000 ep. & \textbf{44.2} & 42.7 & 67.8 \\
\bottomrule
\end{tabular}
\tablenote{Bold compares Mask R-CNN and SAM~3. 3DSG oracle: per-pixel labels rendered from the 3DSG annotation.}
\end{table}

\begin{figure*}[t]
  \centering
  \includegraphics[width=\textwidth]{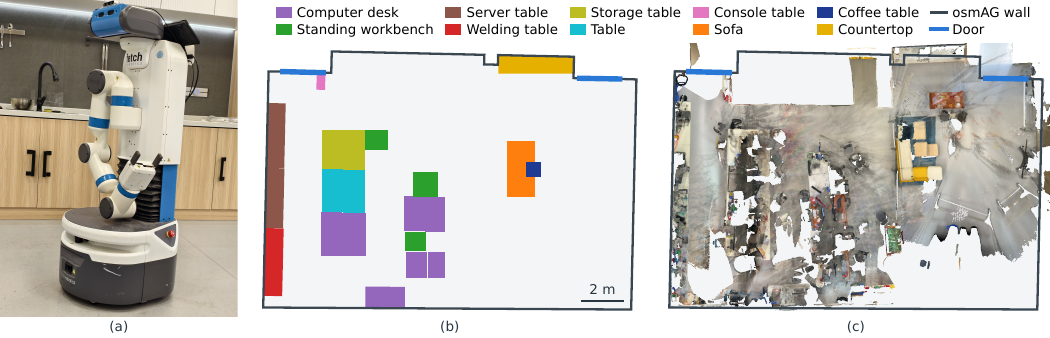}
    \vspace{-8mm}
  \caption{Real-world test scene, a laboratory of over 200\,m$^2$. (a)~The Fetch robot that used for real-world experiments, photographed in the test scene. (b)~The room's osmAG map~\cite{feng2023osmag} with the 21 annotated furniture boxes, the only semantic input, colored by category. (c)~A 3D scan of the same room with the osmAG walls overlaid.}
  \label{fig:real_scene}
\end{figure*}

\subsection{Room-to-Floor Transfer and Efficiency}
\label{sec:tsp_diffusion}

We evaluate whether a search-reference generator trained on room-level data transfers to floor-level scenes while retaining navigation performance close to its teacher. Diffusion is trained only on ScanNet++ and applied directly to Gibson, without training or fine-tuning on Gibson. We compare it with the TSP teacher planner solving the Gibson instances again. Both use the same full-scene furniture prior, and six-category protocol with Mask R-CNN. TSP uses $K=3$ candidates per neighborhood and a 300\,s solver time limit.

\textbf{Approaching teacher performance at lower generation cost.}
With $25.8\times$ less generation time than TSP on average ($165\times$ less at P95), diffusion achieves only 0.7\% lower SR and even gains 0.2\% in SPL with Mask R-CNN (Table~\ref{tab:tsp_diffusion}); with the 3DSG oracle, it drops 0.2\% in SR and gains 2.6\% in SPL. When transferring our diffusion model from the room level to the floor level, navigation performance drops by less than 1\% compared to our TSP teacher, yet SPL actually improves. More importantly, we avoid the high computational costs associated with solvers, and our model's execution time does not increase with the number of furniture items, making it far more feasible for real-world deployment.

\begin{figure}[!t]
  \centering
  \includegraphics[width=\columnwidth]{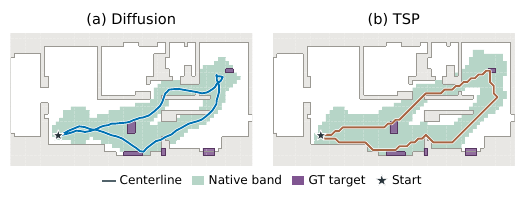}\par\vspace{2pt}
  \includegraphics[width=\columnwidth]{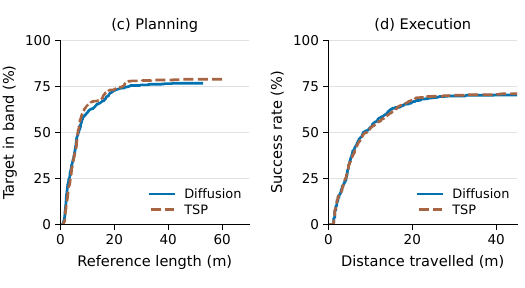}
  \caption{Search references of diffusion and its TSP teacher on the 600 six-category episodes. (a,b)~Search references of diffusion and TSP for one episode in the Gibson validation scene \scene{Darden}, with the query \emph{potted plant}. The native band is the region swept by each generator's own half-widths $w_j$ along its centerline; the purple GT targets are the potted plants in the scene, overlaid for evaluation only. (c)~Planning: share of episodes whose band, swept along the reference, contains the target, against reference length; each curve ends with that generator's longest reference. (d)~Execution: share of episodes that succeed within a distance traveled, under Mask R-CNN.}
  \label{fig:guide_coverage}
\end{figure}

\textbf{Planning: the reference covers the target.}
We first examine the references themselves, on the raw generator output. Let $T_q$ denote the region of the queried category in the top-down GT semantic map, projected onto the planning grid, shown as the GT target in Fig.~\ref{fig:guide_coverage}(a,b), where it comprises all potted plants in the scene; it enters only for evaluation. Sweeping each band along its centerline, we record the reference length at which the band first contains a wall-free cell of $T_q$ (Fig.~\ref{fig:guide_coverage}(c)). By the end of the reference, the band contains the target in 76.7\% of episodes for diffusion and 78.8\% for TSP, and the two curves stay close over the first 20\,m, within which most of this coverage is reached. The learned references are also shorter, 23.1\,m against 30.4\,m on average and 52.4\,m against 61.1\,m at most, where each curve ends.

\textbf{Execution: success per distance traveled.}
For a travel budget $d$, we count the episodes that succeed within $d$ meters of travel under Mask R-CNN (Fig.~\ref{fig:guide_coverage}(d)). The two curves are almost identical: both reach 50\% success before 10\,m, and about 95\% of successful episodes finish within 20\,m, matching the planning curves in Fig.~\ref{fig:guide_coverage}(c). A reference that covers the target thus also lets the agent find it in execution.

\begin{table}[!t]
\centering
\small
\caption{Search-reference generation cost and navigation performance comparison}
\label{tab:tsp_diffusion}
\setlength{\tabcolsep}{2.5pt}
\scriptsize
\begin{tabular}{@{}L{1.15cm} C{0.92cm} C{1.02cm} C{0.72cm} C{0.72cm} C{0.72cm} C{0.72cm}@{}}
\toprule
& \multicolumn{2}{c}{Ref. gen. (s)} & \multicolumn{2}{c}{Mask R-CNN} & \multicolumn{2}{c}{3DSG oracle} \\
\cmidrule(lr){2-3}\cmidrule(lr){4-5}\cmidrule(lr){6-7}
Method & Mean \(\downarrow\) & P95 \(\downarrow\) & SR \(\uparrow\) & SPL \(\uparrow\) & SR \(\uparrow\) & SPL \(\uparrow\) \\
\midrule
TSP & 25.705 & 169.893 & \textbf{71.0} & 39.6 & \textbf{82.7} & 44.0 \\
Diffusion & \textbf{0.995} & \textbf{1.029} & 70.3 & \textbf{39.8} & 82.5 & \textbf{46.6} \\
\bottomrule
\end{tabular}
\end{table}

\begin{figure}[!t]
  \centering
  \includegraphics[width=\columnwidth]{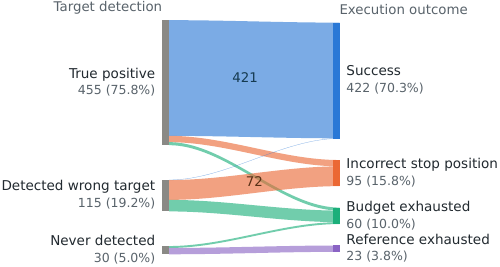}
  \caption{Target recognition and navigation outcomes across 600 episodes. Node labels give counts and percentages; ribbon widths represent episode counts.}
  \label{fig:failure_modes}
\end{figure}

\subsection{Failure Analysis}
\label{sec:failure}
This section addresses the following question: \textbf{What limits navigation success when using the generated search reference?} We analyze the 600 episodes of \method{} on the six-category subset with the full-scene furniture prior and Mask R-CNN (70.3\% SR, Table~\ref{tab:main}), grouping each episode by target detection and execution outcome (Fig.~\ref{fig:failure_modes}). For target detection, we back-project each detection with depth onto the 2D map and take the agent's last estimate of the target position, computing its minimum distance to $T_q$, the queried category's region in the top-down GT semantic map (Sec.~\ref{sec:tsp_diffusion}). A distance of at most 1\,m counts as a \emph{true positive}. The \emph{detected wrong target} group contains the remaining detections: either the estimated position lies more than 1\,m from $T_q$, or no position could be estimated, for example because the detection mask contained no valid depth. For execution, an episode succeeds, stops at an incorrect position, exhausts its 500 action budget, or exhausts the reference without a detection. Success follows almost entirely from a correct target estimate: 421 of the 422 successful episodes are true positives.

Failures, in turn, come mainly from target detection. The detected-wrong-target group (115 episodes, 19.2\%) is the largest source, and its main flow, 72 episodes, ends at an incorrect stop position, where the agent stops at an object mistaken for the target. The few true positives that fail do so in the final stop. Failures traceable to the search reference lie within the never-detected group, only 30 episodes (5.0\%), 23 of which (3.8\%) complete the reference without a detection. Target detection, rather than the search reference, therefore limits navigation success.

\subsection{Real-World and Open-Vocabulary Evaluation}
\label{sec:real}

We deploy \method{} on a Fetch mobile robot with a head-mounted Intel RealSense D435i RGB-D camera, using SAM~3~\cite{carion2025sam3} as the open-vocabulary detector. The test scene is a cluttered laboratory of over 200\,m$^2$, half office area and half leisure area (Fig.~\ref{fig:real_scene}). Its walls come from an osmAG map~\cite{feng2023osmag} aligned to the robot's occupancy grid, and its 21 furniture pieces are annotated once as boxes; the positions of the queried objects are never given.

For each query, the LLM scores over the annotated furniture give $D_q$, and the generator produces the search reference as in simulation. Execution differs in one aspect: instead of following waypoints along the reference, the robot samples viewpoints near the centerline to cover where the band overlaps furniture whose weight $\alpha_c(q)$ clears a threshold, and navigates to them one by one until the target is detected and approached. The prior and the generator are unchanged.

The queries include \emph{tray with fruits}, \emph{robot arm} and \emph{quadruped robot}, which lie outside the COCO vocabulary of the simulation detector. \method{} succeeds in 13 of 15 trials. One failure mode is an object placed far from any furniture: a \emph{quadruped robot} left on the floor lay outside the covered region, so none of the viewpoints brought it into view.

\section{Conclusion}

We presented \method{}, which decides where to search before the robot moves: a conditional diffusion model generates one global search reference from persistent structure and an LLM-derived furniture prior. Under matched structural priors it outperforms SGM on Gibson; trained only on room-level ScanNet++, it transfers zero-shot to floor-level scenes close to its TSP teacher at a fraction of the generation cost, and it demonstrates performance on real-robot with open-vocabulary queries.

\method{} assumes that the target lies on or near furniture. This holds for most everyday objects, but an object left far from any furniture falls outside the band and is not found, as happened with a quadruped robot lying on the open floor of our lab. A natural extension is to search the free space that the reference leaves uncovered when the reference is exhausted.

\bibliographystyle{IEEEtran}
\bibliography{root}

\end{document}